\documentclass[acmlarge]{acmart}
\AtBeginDocument{%
  \providecommand\BibTeX{{%
    \normalfont B\kern-0.5em{\scshape i\kern-0.25em b}\kern-0.8em\TeX}}}
    
\usepackage{graphicx}
\usepackage{amsmath} 
\usepackage{hyperref}
\usepackage{float}
\usepackage{setspace}
\usepackage{fancyhdr}
\usepackage{multirow}
\usepackage{caption}
\usepackage{dirtytalk}
\usepackage[T1]{fontenc}

\acmJournal{JOCCH}

\begin{document}

\title{Syllabification of the Divine Comedy}

\author{Andrea Asperti}
\affiliation{%
  \institution{University  of Bologna,
Department of Informatics:
Science and Engineering (DISI)}
  \streetaddress{Mura Anteo Zamboni 7}
  \city{Bologna}
  \country{Italy}}
   \email{andrea.asperti@unibo.it}

\author{Stefano Dal Bianco}
\affiliation{%
  \institution{University of Siena, Department of philology and literary criticism}
  \streetaddress{Palazzo San Niccolò, via Roma 56}
  \city{Siena}
  \country{Italy}}
\email{stefano.dalbianco@unisi.it}

\begin{abstract}
We provide a syllabification algorithm for the Divine Comedy using
techniques from probabilistic and constraint programming. We 
particularly focus on the {\em synalephe}, addressed in terms of the 
\say{propensity} of a word to take part in a synalephe with adjacent
words. We jointly provide an online vocabulary containing, for each word, information about its syllabification, the location of the tonic accent, and the aforementioned synalephe propensity, on the left and right sides.
The algorithm is intrinsically nondeterministic, producing different possible syllabifications for each verse, with different likelihoods; metric constraints
relative to accents on the 10th, 4th and 6th syllables are used to 
further reduce the solution space. The most likely syllabification 
is hence returned as output. 
We believe that this work could be a major
milestone for a lot of different investigations. From the point of view
of digital humanities it opens new perspectives on computer assisted analysis of digital sources, comprising automated detection of anomalous and problematic cases, metric clustering of verses and their categorization, or more foundational investigations addressing e.g. the phonetic roles of consonants and vowels. From the point of view of 
text processing and deep learning, information about syllabification and the location of accents opens a wide range of exciting perspectives,
from the possibility of automatic {\em learning} syllabification of 
words and verses, to the improvement of generative models, 
aware of metric issues, and more respectful of the expected musicality. 
\end{abstract}

\fancyfoot[RO,LE]{} 
\maketitle

\section{Introduction}
\label{sec:intro}
The Divina Commedia is a famous narrative poem by Dante Alighieri, and one of the most influential works in European literature. 
It is structured into 
three {\em cantiche} (Inferno, Purgatorio and Paradiso), each one composed by 33 {canti}, plus an initial, introductory canto, traditionally considered as part of the Inferno.

The scheme of verses is the so called {\em terza rima}, based on 
lines of eleven syllables (hendecasyllables) \cite{endecasillabo} structured in {\em tercets}
following the rhyme scheme aba, bcb, cdc, ded, \dots
Each tercet is hence composed by 33 syllables, similarly to the number of canti in each cantica.

Differently from the quantitative verse of ancient classical poetry
(Greek and Latin), the hendecasyllable of medieval and modern poetry is
accentual. The key characteristic is the {\em stress on the tenth syllable};
in the frequent case where the final
word is stressed on the penultimate syllable (parola piana), the previous constraint produces verses of eleven syllables. The verse also has a stress preceding the caesura \cite{cesura, Beltrami86}, on either the sixth or the fourth syllable.
The poet may add additional accents providing rhythmic variations that help to obtain stylistic effects. 

An essential prerequisite to follow the rhythm of the verse \cite{Beltrami81,Beltrami84} is to correctly identify its syllables. For several reasons that we shall 
explain in the following, this is not a trivial operation, at the 
point that one is frequently forced to rely on rhythmic constraints to understand the syllabification intended by the author.

The delicate point is the sequence of vowels, that in Italian, 
similarly to other European languages, can be either pronounced as a single sound (diphthongs/triphthongs) or as separate sounds (hiatus).
The phenomenon of {\em dieresis} consists in modifying the 
pronunciation by reading a diphthong splitting the sounds as in a
hiatus; conversely, we have a {\em syneresis} when a hiatus is joined 
into a single sound. 

A similar phenomenon may happen between adjacent words, respectively ending and starting with vowels.
In such a situation, we have to consider that the phono-syntactic chain in natural languages, including Italian, is continuous. When we speak, we don’t make any pause between words. So two vowels facing each other from the borders of two different words normally behave just as if they belong to the same syllable. There is no hiatus. The metric formalization of this phenomenon is called synalephe. According to the structure of the language \cite{Pazzaglia74,Bertinetto78,Bertinetto81}, in Italian poetry synalephe is the norm, providing a sort of “default” way of reading a verse. However, there are many interesting exceptions, both in the perspective of the natural language and of the “poetic licence”, which in Dante’s case is supposed to be almost always well motivated, and aimed to obtain special effects according to the rhythm of the verse \cite{Casella24,Fasani92,Menichetti84,Menichetti93}. The anomalous hiatus between two words is called {\em dialephe}.


For all this reasons, automatic syllabification of Italian poetry, 
and of Dante's work in particular, has been traditionally considered 
as a particularly challenging, almost hopeless task. 
We address the problem using techniques
from probabilistic and constraint programming to generate the most
likelihood syllabification of each verse, delegating to experts the
solution of the most controversial cases. In the current version, we are left with a dozen problematic cases, discussed in Appendix \ref{appendix:anomalous}, whose complexity is not due to syllabification but to the their anomalous rhythm. The automatic identification of
these verses is in fact one of the possible applications of our 
algorithm.

We believe this work may provide a base for a lot of interesting 
investigations. From the point of view of Neural Networks and Deep
Learning, it would be interesting to see if we can teach an agent 
to correctly learn syllabification, and then use this additional 
knowledge to improve automatic generation of verses in 
Dante's style (we shall extensively discuss Poetry generation
in Section~\ref{sec:related}).
Imposing the knowledge of the syllabic structure on a character-base 
textual encoding could provide an essential insight on the role of
vowels, consonants, and their mutual interplay. It would be particularly interesting if we could distinguish (groups of) vowels and consonants
as different clusters in the latent space of their embedding. 

From the linguistic point of view, we could provide automatic
support for the identification of anomalous situations \cite{Casella24}, the identification of periodic functions \cite{Bertinetto73}, 
or the classification of verses into different metric categories \cite{Furioso} (as well as their identification inside large corpora).

While our work is specific for Dante, the overall methodology can be 
easily adapted and fine tuned to any other author\footnote{Some fine tuning seems to be unavoidable, since, e.g. the use of synalephe in Dante may be different from that of Ariosto.}.


\subsection{Achievements}
The syllabification algorithm, developed in the Python language, is 
freely available on GitHub at the address \href{https://github.com/asperti/Dante}{https://github.com/asperti/Dante}. The complete syllabification of the Comedy 
just takes a few seconds.
The current release contains:
\begin{itemize}
\item full syllabification code;
\item a copy of the Comedy's dictionary, saved as a pickle file and 
accessible as a python dictionary data structure. For each word we supply 
information relative to its syllabification (that in a few cases can 
be non deterministic), the position of the accent, and the synalephe 
propensity at left and right extremities;
\item source version of the Divina Commedia, borrowed from the Gutenbgerg
edition (see Section~\ref{sec:gutenberg});
\item syllabified version of the full Comedy, where syllables have been 
divided by a vertical bar as in\smallskip\\
\begin{tabular}{p{1cm}p{7cm}p{4cm}}
&Nel |mez|zo |del |cam|min |di |no|stra |vi|ta, &Inferno I, 1
\end{tabular}
\end{itemize}

\subsection{Structure of the work}
The article is structured in the following way. In Section~\ref{sec:gutenberg}
we mention the digital source that we adopted as a base for our work, as
well as other digital resources available on line, and their related projects.
In Section~\ref{sec:related}, we discuss related works. Since we are not aware
of any other similar syllabification effort, we provide a wider view on the
state of the art in textual processing of poetry, with special emphasis on its generative modeling. In Section~\ref{sec:word-syllabification} we discuss
the syllabification of words, that is intrinsically non deterministic. 
Section~\ref{sec:synalephe} is entirely devoted to our approach to the 
{\em synalephe}, that is the main contribution of this work. 
In Section~\ref{sec:dictionary}, we discuss the structure of the {\em dictionary}, that provides, for each word, information about its syllabification, the location of its accent and the synalephe propensity
of the word with adjacent ones.
In Section~\ref{sec:algorithm}, we outline the syllabification algorithm, 
and our management of the metric constraints. Section~\ref{sec:amendments}
discusses some minor amendments that we did to the Gutenberg edition. A few
problematic cases still remaining to be solved are reported in Section~\ref{sec:problems}.
The final Section~\ref{sec:conclusion} briefly summarizes the main 
contributions of the work, 
discusses possible improvements and hints to the many interesting research
perspectives opened by this work. 

A few addition material is provided in appendices. Appendix~\ref{app:example}
contains, as an example, the full syllabification of the first Canto of the 
Inferno. Appendix~\ref{appendix:hiatuses} contains a detailed investigation
of hiatus/diphthong situations in the Comedy. Appendix~\ref{appendix:anomalous}
provides the list of verses with anomalous rhythm currently identified by
our algorithm.

\section{Digital Source}
\label{sec:gutenberg}
A lot of different digital editions of the Divine Comedy are currently accessible on line; we mention a few of them at the end of this section.
The edition we used for our project was edited by the Project Gutenberg Literary Archive Foundation; specifically, it is the \href{http://www.gutenberg.org/files/1012/1012-0.txt}{ebook n. 1012} of 2015, that is a UTF-8 revision with special 8-bit characters of the version in the ebook n. 1000, based instead on the 7-bit ASCII character set. The Gutenberg
source seems to essentially conform to Petrocchi's critical edition of the
Comedy\footnote{As it is well known, no original manuscript of the Divine Comedy has survived, although there 
are hundreds of ancient manuscripts from the 14th and 15th centuries.
The critical edition edited by Giorgio Petrocchi~\cite{Petrocchi}, published in four volumes between 1966 and 1967 as part of the National Edition of Dante's Works, is usually reputed to be the reference landmark for the Comedy.} \cite{Petrocchi}, used as a reference by most of the digital editions
available on line (see Section \ref{sec:sources}).
The version provides a rich annotation with dieresis and stresses that help the correct syllabification of the text and the disambiguation of words. To make an example, the word Beatrice appears 64 times in the Comedy: in 43 cases 
the couple of vowels \say{ea} constitutes a diphthong, as in\smallskip 

\begin{tabular}{p{1cm}p{6cm}p{4cm}}
&Io son Beatrice, che ti faccio andare, &Inferno II, 70
\end{tabular}\smallskip\\
while in the remaining 21 cases it is a hiatus, as in\smallskip

\begin{tabular}{p{1cm}p{6cm}p{4cm}}
&tra Beatrice e te è questo muro. &Purgatorio XXVII, 36
\end{tabular}\smallskip\\
In the above mentioned Gutenberg edition (as in Petrocchi's one), these 
hiatuses have been marked with a dieresis: Bëatrice.  Not all hiatus/diphthong issues have been solved, however, as we shall see in Section~\ref{sec:word-syllabification}.


As a matter of fact, our tool may provide a valuable assistance in the philological comparison of different editions; for instance we spotted at least one (plausible) error in the Gutenberg version, where there is a difference
with Petrocchi's edition. It is interesting to observe that the mistake is
shared by other digital versions on line, with a possible transfer between 
them: hence, it would have not been revealed by a mere textual comparison.

Apart from a few minor amendments listed in Section~\ref{sec:amendments}, we did a single modification to the Gutenberg text, consisting in replacing some occurrences of single quotes with double ones in a few quotations. 
The reason is that single quotes are identical to apostrophes, that as discussed in section \ref{sec:apostrophe}, play a central role for syllabification; for this reason, it is 
better to avoid any confusion. 

\subsection{Other (re)sources}
\label{sec:sources}
In this section, we mention just a few of the many important projects 
around the world providing digital editions of the Divine Comedy.

\begin{description}
\item[\href{http://dantelab.dartmouth.edu/}{Dante Lab}] \href{http://dantelab.dartmouth.edu/}{http://dantelab.dartmouth.edu/}. This is an on-line database of Dante's commentators, from the 1320s to the present. It offers a virtual workspace supporting the simultaneous visualization and comparison of different texts. Dante Lab was supported by the \href{ttps://dante.dartmouth.edu/}{Dartmouth Dante Project}.

\item[\href{https://dante.princeton.edu/projinfo.html}{Princeton Dante Project}] \href{https://dante.princeton.edu/projinfo.html}{https://dante.princeton.edu/projinfo.html} It offers a richly annotated electronic text for instructional and scholarly use.  The digital source, freely available on line, is based on the critical edition by Giorgio Petrocchi~\cite{Petrocchi}.

\item[\href{https://digitaldante.columbia.edu/dante/divine-comedy/}{Digital Dante}] \href{https://digitaldante.columbia.edu/dante/divine-comedy/}{https://digitaldante.columbia.edu/dante/divine-comedy/} \\
The project results from a collaboration among the Department of Italian, Columbia University Libraries, and Columbia University Libraries' Humanities and History Division. One of its distinctive features is an original way to read and research intertextual passages in the Commedia called Intertextual Dante~\cite{intertextualdante}.

\item[\href{https://www.dantenetwork.it/}{Dante Network}]\href{https://www.dantenetwork.it/}{https://www.dantenetwork.it/}]\\
Dante Network is a platform hosted by the University of Pisa collecting data and tools for the investigation, the enrichment and the enhancement of works by Dante Alighieri. The most recent and innovative
contribution is the Hypermedia Dante Network (HDN), which aims to 
extend to the Divina Commedia the ontology and the tools already 
tested on minor works of Dante Alighieri.

\item[Wikisource] Wikisource offers a couple of versions of the Divina Commedia; one based on Petrocchi's edition (\href{https://it.wikisource.org/wiki/Divina_Commedia}{https://it.wikisource.org/wiki/Divina\_Commedia}),  and another one in the edition of Francesco da Buti\footnote{Francesco da Buti, also known as Francesco di Bartolo (Pisa or Buti, 1324 - Pisa, 25 July 1406 ), was an Italian literary critic 
and Latinist, and one of the first commentators on the Divine Comedy.} (\href{https://it.wikisource.org/wiki/Commedia_(Buti)}{https://it.wikisource.org/wiki/Commedia\_(Buti)} .

\item[\href{https://www.treccani.it/enciclopedia/elenco-opere/Enciclopedia_Dantesca}{Enciclopedia Dantesca Treccani}] \href{https://www.treccani.it/enciclopedia/elenco-opere/Enciclopedia_Dantesca}{https://www.treccani.it/enciclopedia/elenco-opere/Enciclopedia\_Dantesca}\\ This enciclopedia, freely accessible on line, provides valuable information about Dante's vocabulary, 
and was a relevant source in our work.

\end{description}

\section{Related Works}
\label{sec:related}
A pioneering application of computers to metric investigations of the Divina Commedia can be found in \cite{Bertinetto73},
aiming at a statistical analysis of the so called {\em periodic functions}, that is
the identification of recurrent patterns in the position of accents inside
the verse. The study, implemented in FORTRAN on a IBM 360/44, 
was performed on two subsets of 1272 verses (dataset A) and 1024 verses (dataset
B) extracted according to different policies from the Comedy. 
The verses in these two subsets were {\em manually preprocessed} to split 
them into {\em metric units} of suitable length, as e.g.
\[1/4/2/2/2/\]
corresponding to accents at the following positions:
\[-+---+-+-+-\]
More recently, a similar investigation has been done by the second author
in the case of the {\em Orlando Furioso} by Ludovico Ariosto \cite{Furioso}.

However, at the best of our knowledge, and in spite of the many projects in digital humanities focused on Dante's work and the Divina Commedia, some of which have been recalled in Section~\ref{sec:sources}, no one addressed so far the issue of automatic syllabification, nor provided as we do, a complete syllabification of the Comedy. 

This is not at all trivial, due to the complex nature
of the Italian hendecasyllable, and the sophisticated interplay between
metric accents and synalephe. Since the objective of our study is original, our methodology and techniques are original too, and essentially developed from scratch. 

Similarly to the field of digital humanities, there is a lot of interest, at present, on textual processing of poetic literature in the field of Deep 
Learning. Here, the final aim is the development of good generative textual 
models; there is a wide perception that the additional challenges posed by meter and rhyme could possibly
drive the discovery of new techniques, fostering the development of the field. Since the seminal work of A.Karpathy~\cite{KarpathyL15}, exploiting Recursive Neural Network for automatic textual 
generation of image captions,
(see also his wondrous blog  \href{http://karpathy.github.io/2015/05/21/rnn-effectiveness/}{\say{The Unreasonable  
Effectiveness of Recurrent Neural Networks}}, where RNN are applied for the
first time to Shakespeare's work) there has been a lot of work on Deep 
Learning techniques for Natural Language Processing (NLP), and also a specific interest on poetry, 
in many different languages, comprising e.g. English \cite{deepspeare},
Chinese \cite{Li-Bai}, ancient Greek \cite{Homer}, and also Italian
\cite{Zugarini}. An additional source of complexity is that the problem can be addressed at different linguistic levels:
characters, syllables, sub-words, or words. Moreover, the very 
notion of rhyme, in the Italian language, cover the last part of the word from the tonic accent, requiring knowledge about its position,
and possibly ad-hoc tokenization. Works on poetry generation frequently rely on specific decoding
procedures to generate reasonable poetry, involving selection from a set of candidate outputs \cite{survey,Ghazvininejad,Zugarini}. Rhyming and metric constraints are somehow orthogonal to each
other and can also be decoupled: working on the final word of verses, it is relatively easy to learn rhyming patterns relying e.g. on similarity matrix between words \cite{RhymingConstraints}.

While rhyme is usually reputed to be the main problem of poetry generation, in the case of the Italian
endecasyllable, metric constraints are the real challenge. As we shall explain, the musicality \cite{Bianchi25} of
the verse is related to the position of stresses inside the verse, whose actual location, however, 
can be at some extent modified by interpretative modulations, via dieresis and dialephe. This is 
precisely what makes syllabification hard.

The first author recently proposed automatic generation of verses in Dante's style as a student project for his course of Deep Learning at the University of Bologna.
In spite of the state-of-the art techniques exploited by students, 
comprising word embeddings \cite{Word2vec}, attention \cite{attention,Multi-head}, transformers \cite{transformer} (replacing RNN
in most NLP applications), including the \say{light} version (117 millions
parameters) of GPT2 \cite{GPT2}, results remained modest. The respect of endecasyllables
is erratic, and understanding the rhyme structure is still problematic (jointly learning the {\em what} and the {\em where} still seems to be an issue). There are many justifications for these modest results, starting
from the relatively small dimension of the training set. However, 
we believe that, in the case of Italian poetry, an additional source of
problems is the lack of sufficient information in the source data, that
is one of the motivations for decorating the text with the correct 
syllabification, and provide the position of the accent inside the word.
This enriched information could also, at some extent, remedy to the 
limited dimension of the training set.

Independently from the generative task, our enriched format opens the way
to a lot of interesting investigations, already hinted to in the 
introduction, especially in relation with a combined processing 
of the text at character, syllable, and word level, aimed to investigate
their mutual roles.


\section{Syllabification of words}
\label{sec:word-syllabification}
As we already recalled, the main problem of syllabification concerns the division of groups of vowels. Specifically, there are two main subcases, depending on whether the vowels are intra-word (hiatus/diphthong) or inter-word (synalephe). The synalephe is by far the most interesting and complex phenomenon, and the  main subject of this research; it will be discussed in the next section. 
Our approach presupposes, among other things, a known, possibly non deterministic, syllabification of words, that we shall address in this 
Section.

The syllabification of single words, even if different from that of modern Italian, is not a really compelling problem; however, several words
require ad-hoc treatment, and the situation is further complicated by the
frequent occurrences of Latin words, and words that, in Dante's style, recall the spelling of Latin language, not to speak of the famous Provencal
tercets of Arnaut Daniel (Arnaldo Daniello) in Purgatorio XIV, 140-147.

For all these reasons, we do not release a syllabification code, but instead provide a {\em full dictionary} with the syllabification of {\em each word} occurring in the Divine Comedy. This syllabification is {\em non-deterministic}, due to the fact that a single word may be split in different ways in different context, for metric reasons. 
A typical example, that we already discussed, is the word \say{Beatrice}.

A deeper investigation of hiatuses and diphthongs in the Divine
Comedy is given in Appendix \ref{appendix:hiatuses}. 
In this section, we briefly discuss a few interesting issues, that may 
also help to understand the facilities offered by our tool for the analysis of the document.

As we already remarked, most of the hiatus/diphthong dichotomies are solved
if the Gutenberg edition, by the use of dieresis. 
However, 
\begin{enumerate}
\item not all problems have been addressed in the text;
\item the use of dieresis may be questionable.
\end{enumerate}

Concerning the latter issue, we shall discuss a few delicate cases
in Appendix \ref{appendix:hiatuses}, mostly relative to the groups of vowels 
\say{io} and \say{ea}.

Here we discuss a couple of the remaining hiatus/diphthong problems 
still present, in our opinion, in the Gutenberg edition. In all these
cases, instead of superimposing notation on the text, e.g.
in the form of additional dieresis, we exploit the intrinsic {\em non
determinism} of our approach, allowing {\em multiple syllabifications} 
to draw upon in a probabilistic way. 
It is also important to observe that
all these problematic samples have been {\em automatically} evinced by our algorithm, since syllabification failed or was extremely unlikely.

A first interesting case is relative to the word \say{creature} (creatures).
Such word appears 8 times in the Divine Comedy; 
invariably, the sequence of vowels \say{ea} is a hiatus, but for
a single unfortunate verse in the Paradiso:\smallskip

\begin{tabular}{p{1cm}p{6cm}p{4cm}}
&e queste cose pur furon creature;  & Paradiso VII, 127
\end{tabular}\smallskip

Unfortunately, there is no accepted notation to express syneresis, so it
is convenient to address the problem allowing a double, nondeterministic syllabification for the word \say{creature}, with sensibly different 
probabilities.

Some other interesting cases are the words ending with -aio (primaio, Tegghiaio\footnote{Tegghiaio Aldobrandi was mayor of San Gimignano and Arezzo; he fought in the battle of Montaperti (1260) as a Guelph.}, migliaio and similar), which are normally spelt with a hiatus “a-io”:\smallskip

\begin{tabular}{p{1cm}p{6cm}p{4cm}}
&E questi sette col primaio stuolo  &Purgatorio XXIX, 145\smallskip\\
&migliaia di lunari hanno punita.  &Purgatorio XXII, 36
\end{tabular}\smallskip

However, for an ancient habit in Italian poetry\footnote{A habit which probably deals – by extension – with the monosyllabic treatment of the word \say{gioia} in Italian manuscripts, coming from the key-word \say{joi} in provencal poetry \cite{Menichetti93}, pp. 293-4},  the same suffix can appear as monosyllabic:\smallskip

\begin{tabular}{p{1cm}p{6cm}p{4cm}}
&Farinata e ’l Tegghiaio, che fuor sì degni &Inferno VI, 79\smallskip\\
&Quanto di qua per un migliaio si conta &Purgatorio XIII, 22\smallskip\\
&ne lo stato primaio non si rinselva. &Purgatorio, canto XIV, 66
\end{tabular}



%


\section{Synalephe}
\label{sec:synalephe}
The main problem for the correct syllabification of the Divine Comedy is the synalephe, that is the \say{melding} into a single syllable of two vowels belonging to two different, adjacent words \cite{cesura,Beltrami84,Bertinetto81,Menichetti93}.


Our approach to the synalephe consists in associating to each extremity of a word
a {\em probability}, that can be understood as its propensity for taking part into
a synalephe. This probability can be computed statistically, computing the frequency 
of the phenomenon. 

Since the other important metric information associated to a word is the number 
$n$ of its syllables, each word can be schematically described as a triple
\[\langle p_l, n, p_r \rangle\] 
where $p_l$ and $p_r$ are the synalephe probabilities (left, and right, respectively)
and $n$ is the number of syllables of the word (we will be forced to extend this basic representation with additional information). 

The general idea is that the probability to have a synalephe between two adjacent 
words $w_1 = \langle p^1_l, n^1, p^1_r \rangle$ and $w_2=\langle p^2_l, n^2, p^2_r \rangle$ is given by the product $p^1_r p^2_l$.

For words starting/ending with a vowel, the (left/right) synalephe is the norm. So, in typical cases, the synalephe probability is $1$ for vowels and $0$ for consonants. 
For instance, the word \say{selva} is described by the triple 
$\langle 0, 2, 1 \rangle$, since it starts with a consonant, it has 2 syllables and
it ends with a vowel; similarly, \say{oscura} is associated with the triple 
$\langle 1, 3, 1 \rangle$ since it starts and ends with a vowel, and it contains 3 syllables.

\begin{center}
\begin{tabular}{cc}
selva & oscura\\
$\langle 0, 2, 1 \rangle$ & $\langle 1, 3, 1 \rangle$
\end{tabular}
\end{center}
When we compose the two words together, the probability to have a synalephe is
$p^1_r p^2_l = 1$. Since we have a synalephe, in the computation of the total
number of syllables we need to subtract 1 to the sum of $n_1$ and $n_2$, In the 
above case, we have $2+3-1=4$.

When the probability of having a synalephe is neither 0 nor 1, we need to consider
both possibilities, with the corresponding probabilities. In the end, we shall 
choose the most likelihood syllabification for the verse, compatible with 
the metric constraints. In principle, the number of cases may grow 
exponentially with the length of the verse. In practice, this is never a problem, 
since (a) relatively few words have not-categorical synalephe probabilities and 
(b) the length of the verse is in any case very small.

In our dictionary of the Divine Comedy we provide left and right estimations of the synalephe probabilities for each word. 

The previous algorithm can be easily extended to take into account multiple, non deterministic syllabification of single words. For instance, we might be interested to 
associate to the word \say{avea} two possible representations, corresponding to 
the case of dipththong/hiatus (where the latter should have precedence on the former,
if compatible with the metric):
\begin{center}
\begin{tabular}{ccc}
& avea &\\
$\swarrow\hspace{-.4cm}$ & &\hspace{-.4cm}$\searrow$\\
$\langle 1, 2, 0.1 \rangle$& & $\langle 1, 3, 1 \rangle$
\end{tabular}
\end{center}
In the case of diphthong, the words contains 2 syllables, and the right synalephe is unlikely; in the case of hiatus, the word contains 3 syllables, with definite propensity for synalephe. In the case \say{avea} is followed by another word starting with a consonant, the hiatus version will typically generate verses incompatible with the metric constraints, and the diphthong version will eventually
prevail\footnote{In the current version, we do not follow this approach for the word \say{avea}. We
systematically treat it as a diphthong, unless if otherwise required 
by the editor with a dieresis annotation.}.

\subsection{Metric constraints}
\label{sec:metric-tuples}
The main metric constraint is to have a stress on the tenth syllable. 
As a rough approximation, we could expect 11 syllables for each verse, 
but a more precise investigation eventually requires to take into account
the position of the stress inside the word. This is not a completely trivial
operation in case of Dante, due to the old lexicon, the sometimes unusual 
conjugation of verbs and other problems. However, in our dictionary we
also provide this auxiliary information. The \say{triple} of the previous
section now becomes a quadruple
\[\langle p_l,n,a,p_r \rangle\]
where the new information $a$ is an integer in the range [-(n-1),0] expressing the position of the accent as a negative offset from the right. For example, in 
\say{carità} the offset is 0, and in \say{Ettore} the offset is -2. 

The main advantage of taking accents into consideration is the fact that, in 
addition to the stress on the tenth syllable, we can also add other metric 
constraints, for instance relative to the stress preceding the caesura, 
on either the sixth or the fourth syllable (or more sophisticated patterns).
Syllabifications not satisfying these constraints can be pruned or 
severely penalized in their likelihood.

There is just a slight difficulty in expressing the constraint 
relative to the stress on the tenth syllable, regarding what we may
expect {\em after} the stress. If the stress is on the last word, we accept
the verse even if it has more than 11 syllables, to take into account
the few {\em versi sdruccioli} of the Comedy, such as:\smallskip\smallskip

\begin{tabular}{p{.4cm}p{6.6cm}p{4cm}}
&che noi possiam ne l’altra bolgia scendere,  &Inferno XXIII, 32\smallskip\\
&ch’era ronchioso, stretto e malagevole,  &Inferno XXIV, 62\smallskip\\
&non da pirate, non da gente argolica.  &Inferno XXVIII, 84\smallskip
\end{tabular}\smallskip

In general, we would be tempted to refuse any additional word after the one 
containing the stress in $10^{th}$, but this would rule out a few interesting
cases of {\em composed rhyme} \cite{Beltrami84,Menichetti93}. For instance, in the following verse\smallskip

\begin{tabular}{p{.4cm}p{6.6cm}p{4cm}}
&e men d’un mezzo di traverso non ci ha &Inferno XXX, 87
\end{tabular}\smallskip\\
the stress is on \say{non} and nevertheless it is followed by {\em two} additional words (counting as one, due to the synalephe between \say{ci} and \say{ha}). 
If you are curious, the verse is rhyming with \say{oncia} and
\say{sconcia} (sic!).

Another example is\smallskip

\begin{tabular}{p{.4cm}p{6.6cm}p{4cm}}
&Poi che ciascuno fu tornato ne lo  &Paradiso XI, 13
\end{tabular}\smallskip\\
(ryhiming with \say{cielo} and \say{candelo}). The accent is on 
\say{ne} (a bland proposition, by the way), followed by an additional
word. 

The rule we adopt is that, after the word containing the accent on the
tenth syllable (that may have arbitrary length), we may accept additional words provided they do not trespass the total amount of 11 syllables. 

\subsection{The role of punctuation}\label{sec:punctuation}
As in all known metric systems, the phono-syntactic continuity is not affected by punctuation marks. The presence of a punctuation mark is therefore substantially irrelevant for the purposes of the syllabification of the verse. 

Normally, even the dot does not prevent synalephe:\smallskip\\
\begin{tabular}{p{1cm}p{7cm}p{4cm}}
&per simil colpa». E più non fé parola.  &Inferno VI, 57\smallskip\\
&fossero». Ed ei mi disse: «Il foco etterno  &Inferno VIII, 73\smallskip\\
&perduto». Ed elli: «Vedi ch’a ciò penso».  &Inferno XI, 15\smallskip\\
&sotto la pece?». E quelli: «I’ mi partii,  &Inferno XXII, 66\smallskip\\
&diss’ io, «chi siete?». E quei piegaro i colli;  &Inferno XXXII, 44
\end{tabular}\smallskip

As a matter of fact, punctuation marks are an acquisition of modern philology, from the Renaissance onward, when the first publishers of printed texts faced the problem of how to reproduce the text of ancient manuscripts, where, as it is well known, punctuation is absent,
or reduced to the essential.

\subsection{Diving deeper into the synalephe}
\label{sec:deeper}
The idea of associating a synalephe probability 1 to vowels and 0 to
consonants is of course just a basic, rough approximation. There are many
exception to this rule and then exceptions to exceptions, leading us
naturally to a probabilistic approach. 

We now investigate a few subcategories of words deserving a special 
treatment. \smallskip

{\bf Remark}
It is important to remark that the following discussion is {\em not} intended to be an exhaustive analysis of all cases of synalehpe/dialephe inside
the Comedy. It is only meant to drive a reasonable {\em initialization} of the
synalephe probabilities, ruling out some relatively trivial cases. The algorithm will automatically spot the remaining problematic cases, for which probabilities
can be assigned in an automatic or supervised way.\smallskip

\subsubsection{Words ending in an accented vowel}
In the vast majority of cases, an accented final vowel requires
dialephe. The words in the following, relatively short, list appear in the Comedy both with dialephe and synalephe (or exclusively with synalephe\footnote{likely due to the small number of occurrences.}):\smallskip\smallskip

apparì, bontà, ché, drizzò, fé, già, là, lì, lasciò, perché, però, più, portò, ricominciò, sé, sì, tornò, turbò\smallskip\smallskip

\noindent
Here are some examples (the first one with synalephe, and the second one 
with dialephe)\smallskip

portò
 
\begin{tabular}{p{.4cm}p{6.6cm}p{4cm}}
&sì che, stracciando, ne portò un lacerto. &Inferno XXII, 72\smallskip\\
&A Minòs mi portò; e quelli attorse  &Inferno XXVII, 124
\end{tabular}\smallskip

perché

\begin{tabular}{p{.4cm}p{6.6cm}p{4cm}}
&perché appressando sé al suo disire,  &Paradiso I, 7\smallskip\\
&perché ardire e franchezza non hai,   &Inferno II, 123
\end{tabular}\smallskip

tornò

\begin{tabular}{p{.4cm}p{6.6cm}p{4cm}}
&Noi ci allegrammo, e tosto tornò in pianto;  &Inferno XXVI, 136\smallskip\\
&Così tornò, e più non volle udirmi.  &Purgatorio XVI, 145
\end{tabular}\smallskip

\subsubsection{Words ending with a diphthong}
In Dante, groups of vowels at the end of words are frequently treated as 
diphthongs/triphthongs. For instance: \smallskip

\begin{tabular}{p{.4cm}p{6.6cm}p{4cm}}
&al pel del vermo r\textbf{eo} che ’l mondo fóra.  &Inferno XXXIV, 108\smallskip\\
&del gran dis\textbf{io}, di retro a quel condotto  &Purgatorio IV, 9\smallskip\\
&che pr\textbf{ia} m’av\textbf{ea} parlato, sorridendo  &Paradiso XI, 17\smallskip\\
&ch’uscir dov\textbf{ea} di l\textbf{ui} e de le rede;  &Paradiso XII, 66\smallskip\\
\end{tabular}\smallskip

When a word with a potential hiatus is followed by another one starting with a vowel there is an ambiguous situation. For instance,
in the verse \smallskip

\begin{tabular}{p{.4cm}p{6.6cm}p{4cm}}
&tien alto lor disio e nol nasconde.  &Purgatorio XXIV, 111
\end{tabular}\smallskip\\
we could split \say{disio e} as \say{di|sio |e} or 
\say{di|si|o e}.\\
Menichetti \cite{Menichetti93} treats these cases and calls them {\em diesinalefe}: when the possibility of a dieresis meets the possibility of a dialephe, the dialephe wins.


We privilege the diphthong also at the end of verses. So, a verse like\smallskip 

\begin{tabular}{p{.4cm}p{6.6cm}p{4cm}}
&mentre ch’io vissi, per lo gran disio  &Purgatorio XI, 86
\end{tabular}\smallskip\\
is going to be treated as an hendecasyllable with ten syllables. 

This assumption is somewhat in contrast with the Italian metric tradition, where these cases are treated as {\em endecasillabi piani}, with eleven syllables 
\cite{Pazzaglia74,Beltrami84,Menichetti93}. 

However, the previous choice has essentially no impact on the metric structure of the verse, while it is sensibly simpler from the algorithmic point of view, since the syllabification of words becomes independent from their position inside the verse. If required, these cases are easily recognizable and could be simply fixed in a post processing phase. 

In typical cases, and coherently with the previous politics, words ending with
a diphthong have no propensity to synalephe. 

\subsubsection{words starting with a diphthong}
Similarly to the previous case, when a word start with \say{ia} (Iacopo, iaculi, iattura, \dots), \say{io} (Iosüè, Iove,\dots), \say{ie} (ier, iernotte, \dots), \say{iu} (iura, iube, Iunone, \dots) it is not inclined to (left) synalephe.

Examples are\smallskip

\begin{tabular}{p{.4cm}p{6.6cm}p{4cm}}
&«O Iacopo», dicea, «da Santo Andrea &Inferno XIII, 133\smallskip\\
&ché se chelidri, iaculi e faree  &Inferno XXIV, 86\smallskip\\
&qual diverrebbe Iove, s’elli e Marte & Paradiso XXVII, 14\smallskip\\
&Chi dietro a iura e chi ad amforismi  &Paradiso XI, 4\smallskip\\
\end{tabular}\smallskip\\

Other combinations of vowels are less evident (and less frequent, too).
An interesting case is e.g. the word \say{uomini}, sometimes participating in synalpehe\smallskip

\begin{tabular}{p{.4cm}p{6.6cm}p{4cm}}
&Li uomini poi che ’ntorno erano sparti  &Inferno XX, 88
\end{tabular}\smallskip\\
and sometimes not:\smallskip

\begin{tabular}{p{.4cm}p{6.6cm}p{4cm}}
&Ahi Genovesi, uomini diversi  &Inferno XXXIII, 151
\end{tabular}

\subsubsection{Monosyllables} 
Another class of words deserving a special attention
is that of monosyllables. For instance, it is known \cite{Menichetti93} that a monosyllable
followed by a word starting with an accented vowel typically results in a dialephe.
However, in Dante, monosyllables frequently induce a dialephe even in different
situations, like e.g. \smallskip

\begin{tabular}{p{.4cm}p{6.6cm}p{4cm}}
&a chi avesse quei lumi divini  &Paradiso VIII, 25\smallskip\\
&dico che arrivammo ad una landa  &Inferno XIV, 8\smallskip\\
&che alcuna virtù nostra comprenda,  &Purgatorio IV, 2\smallskip\\
&punto del cerchio in che avanti s’era,  &Paradiso XI, 14\smallskip\\
\end{tabular}\smallskip

It is also interesting to observe that many monosyllables end with with a diphthong
(mio, tuo, suo, noi, voi, poi, cui, fui, sia, via, etc.), or are accented (già, ché, fé, già, là, più, sé, sì, etc.) so they also fall into the previous categories. 

According to our investigation, the following additional monosyllables never occur in a synalephe in the Comedy:

\hspace{1cm}Be, me, fa, fo, mo, Po, pro, qua, re, sto, te, tu, tra, tre

The following monosyllables deserves instead a probabilistic treatment (in addition to
the accented words already listed in the previous section):

\hspace{1cm}a, ad, che, chi, da, e, fra, fu, io, ho, ha, ma, o, qui, se, su, va \hfill(*)

\noindent
For instance, the quadruple associated with the conjunction \say{e} (and), is
\[\langle .9,1,0,.2 \rangle \]
meaning that it has a high propensity to take part in a synalephe on the left and
is available (but not really inclined) to participate in a synalephe on the right. 
So, if there is the possibility to have a synalephe both on the right and on the left, the left one will be eventually preferred. For instance, in the verse\smallskip

\begin{tabular}{p{.4cm}p{6.6cm}p{4cm}}
&che membra feminine avieno e atto,  &Inferno IX, 39
\end{tabular}\smallskip\\
the syllabification would be \say{a|vie|no e|at|to} and not 
\say{a|vie|no| e at|to}.

The synalephe probalities for the words in the (*) list are very
different from each other. For instance, words like \say{da}, \say{ma} or \say{fu} have very small synalephe propensity, that is typical of Dante; quoting Menichetti (translation by the authors):

\begin{quotation}
Dante always has hard attack (and dialephe) after the preposition 
\say{da}. The scarce availability of a particle for elision is in many cases an indication that it is normally followed by a hiatus in the language, without having to be syntagmatically tonic: \say{Ma Amore}, four syllables.\flushright Menichetti \cite{Menichetti93}, p.338
\end{quotation}

In general, the necessity of a probabilistic approach is 
readily understood in view of the following observation by Menichetti (translation by the authors):

\begin{quotation}
The unstressed vowels subject to apocope or elision (\say{di} better than \say{da}, for example) and the initial apheretizable ones (especially i-) […] bear particularly well the synalephe, and it is instead normal that they do more than other obstacles to the dialephe. Dante, who after the syntagmatically unstressed word \say{fu} almost always has dialephe (Par IX, 120), however, has synalephe with i- (Inf IXX, 63, Inf V, 54). Thus the dialephe of Par XX, 38 assumes special importance.\flushright Menichetti \cite{Menichetti93}, p.339
\end{quotation}
Remarkably, all verses mentioned by Menichetti are perfectly syllabified by our algorithm.

\subsection{Apostrophes}\label{sec:apostrophe}
An apostrophe is used to express the elision of a part of a word. Typically,  
the elision refers to a final or initial vowel, but in rare cases it may 
replace an entire syllable, as in \say{ver’ },\say{inver’ } (towards), where the
corresponding full words are \say{verso},\say{inverso}, or also delete letters
inside words, as in \say{acco’lo} (accoilo) or \say{entra’mi} (entraimi). 
These latter cases must be treated
in a special way: in particular, the apostrophe has essentially no effect on 
syllabification. 

In the other cases, the apostrophe should be basically treated as
a vowel. For this reason, we prefer a syllabification of the form \say{do|v’ or} 
over the more grammatical \say{dov’ | or} (again, this can be possibly adjusted, 
if required, in postprocessing). Observe that, with our encoding of words, the two forms
are sensibly different: \say{do|v’ } has two syllables and is definitely prone to
synalephe; \say{dov’ } has a single syllable and refuses synalephe. 

Our management of the apostrophe is however a bit more sophisticated. 
The point is that the apostrophe is not just expressing an elision, but frequently 
it is also meant to {\em induce} a synalephe in situations where, in principle, we would not
expect one.

To make an example, let us consider the following verse:\smallskip\\
\begin{tabular}{p{1cm}p{6.6cm}p{4cm}}
&Così vid’ i’ adunar la bella scola  &Inferno IV, 94
\end{tabular}\smallskip\\
that must be syllabified as follows:\smallskip\\
\begin{tabular}{p{1cm}p{6.6cm}p{4cm}}
&Co|sì |vi|d’ i’ a|du|nar| la| bel|la| sco|la  &Inferno IV, 94
\end{tabular}\smallskip

The delicate point is the synalephe between \say{i’} and \say{adunar}. In Dantes'work, the word \say{io} in
similar situations {\em would not} generate a synalephe: consider for instance\smallskip\\
\begin{tabular}{p{1cm}p{6.6cm}p{4cm}}
&E io a lui: «L’angoscia che tu hai  &Inferno VI, 43\smallskip\\
&E io anima trista non son sola,  &Inferno VI, 55\smallskip\\
&per ch’io avante l’occhio intento sbarro.  &Inferno VIII, 66\smallskip\\
\end{tabular}\smallskip

The following case is symmetric:\smallskip\\
\begin{tabular}{p{1cm}p{6.6cm}p{4cm}}
&per ch’io ’ndugiai al fine i buon sospiri,  &Purgatorio IV, 132
\end{tabular}\smallskip

As another example let us consider the verse\smallskip\\
\begin{tabular}{p{1cm}p{6.6cm}p{4cm}}
&E noi lasciammo lor così ’mpacciati.  &Inferno XXII, 151
\end{tabular}\smallskip\\
to be compared e.g. with\smallskip\\
\begin{tabular}{p{1cm}p{6.6cm}p{4cm}}
&che fu al dire e al far così intero.  &Purgatorio XVII, 30
\end{tabular}\smallskip

For all these reasons, we deserve to the apostrophe a special treatment: in our 
embedding of words as tuples, instead of the synalephe probability, we use the 
numerical value 2 to express the presence of an apostrophe (left or right, respectively):
a sort of super-propensity to synalephe.

\section{The dictionary}
\label{sec:dictionary}
An important contribution of our work is to offer a digital dictionary of the Divine Comedy providing word level syllabification as well
as the position of the stress inside the word. In addition, the dictionary
gives left and right synalephe probabilities for each word, in the sense
explained in Section~\ref{sec:synalephe}. 

More specifically, the information associated with each word $w$ is a 
{\em list} of pairs $(t,ws)$, where t is the metric-tuple $\langle p_l,n,a,p_r \rangle$ introduced in Section~\ref{sec:metric-tuples}, and $ws$ is the corresponding syllabification. We remember that $n$ is the number of
syllables in the word, $a$ is the position of the accent (expressed as a
negative offset from the right) and $p_l,p_r$ are the synalephe probabilities
for the word (left and right, respectively).

Usually the list associated with a word $w$ is composed by a single entry,
but in a few cases we need to take into account the possibility to
have multiple, non-deterministic syllabifications.

The syllabification of words is expressed by a string where syllables are separated by a vertical bar character \say{|}. 

A typical usage of our algorithm consists in overriding an entry for a given word 
in the dictionary for discovering interesting cases of synalephe/dialephe (or also diphthong/hiatus) involving that word inside the Comedy. Most of the examples in this
article have been obtained in that way.

\section{The verse syllabification algorithm}
\label{sec:algorithm}
We process a verse as a sequence of tokens $w_1$, \dots, $w_n$. Tokens
can be words or punctuation symbols; however, as explained in 
Section~\ref{sec:punctuation} punctuation symbols play essentially no role
and can be neglected. 

Remarkably, the only information we need for each word is its metric
tuple $\langle p_l,n,a,p_r \rangle$.

Suppose we already processed some initial part of the verse, up to
token $i$. The information we have (the current {\em state}, in computer 
science terminology), 
is expressed in a list of
possible syllabifications up the current token, where each
syllabification is enriched with information regarding its likelihood
$p$, the current number of syllables, and the synalephe probability of its
last word.  

To make an example, consider the simple verse 

\begin{tabular}{p{1cm}p{7cm}p{4cm}}
&Nel mezzo del cammin di nostra vita.  &Inferno I, 1\smallskip\\
\end{tabular}\smallskip\\
and suppose that we already processed the initial part of the verse up
to, say, the word \say{cammin}. The current {\em state} would be described by
the following tuple
\[
    (\underbrace{\mbox{|Nel |mez|zo |del |cam|min}}_{\mbox{syllabification}},\underbrace{1}_{\mbox{likelihood}},\underbrace{6}_{\mbox{syll. no.}},\underbrace{0}_{\mbox{$pr$}})
\]
The first element is the actual syllabification; the second element is its 
likelihood, that in this case is 1, since it is deterministic (and hence unique), the third element
is the current number of syllables, that in this case is 6, and the final one
is the synalephe propensity of the last word \say{cammin}, that is 0.

If we had multiple possible syllabifications, their likelihood would be distributed among them.

Suppose now we process the next token \say{di}. We just need to take into 
account the possibility of a synalephe. In this case it is $0$ because both
adjacent words exclude it, so we add a syllable separator, concatenate the
syllabification of the word \say{di} (obtained form the dictionary) to the
current syllabification of the verse, update the number of syllables, and
remember the synalephe probability $p_l$ of the last word \say{di}. So, 
the new state will be:
\[(\mbox{|Nel |mez|zo |del |cam|min |di},1,7,?)\]
The probability of having a synalephe between two adjacent words with 
synalephe probabilities $p_r$ and $p_l$ (left and right, respectively) 
is just the product between the two probabilities, namely $p_rp_l$.
In case one of $p_r$ or $p_l$ is 2 (corresponding to an apostrophe, see 
Section~\ref{sec:apostrophe}), than the resulting probability is 1, independently from the other.

If the probability $p$ of having a synalephe is not categorical (0 or 1), 
we need to consider both possibilities, where the respective likelihood
is the product between the current verse likelihood and the probability
of the case under consideration ($p$ for synalephe and $1-p$ for absence of
synalephe).

Let us discuss a slightly more complex example, considering the verse\smallskip\\ 
\begin{tabular}{p{1cm}p{6cm}p{4cm}}
&esta selva selvaggia e aspra e forte  &Inferno I, 5
\end{tabular}\smallskip\\
We have no problems up to the first \say{e}, where the state is
\[(\mbox{|e|sta |sel|va |sel|vag|gia},1,7,1)\]
Remember that the tuple for the word \say{e} is $\langle .9,1,0,.2$.
So, after processing \say{e} we have two possibilities:
\[
\begin{array}{l}
(\mbox{|e|sta |sel|va |sel|vag|gia e},.9,7,.2)\\
(\mbox{|e|sta |sel|va |sel|vag|gia |e},.1,8,.2)\\
\end{array}
\]
The left synalephe probability for \say{aspra} is 1, so, for both states,
the probability to have a synalephe is $.2$. After processing \say{aspra}
we end up with 4 possible states, listed according to their probabilities
\[
\begin{array}{l}
(\mbox{|e|sta |sel|va |sel|vag|gia e |a|spra},.72,9,1)\\
(\mbox{|e|sta |sel|va |sel|vag|gia e a|spra},.18,8,1)\\
(\mbox{|e|sta |sel|va |sel|vag|gia |e |a|spra},.08,10,1)\\
(\mbox{|e|sta |sel|va |sel|vag|gia |e a|spra},.02,8,1)\\
\end{array}
\]
After processing the next \say{e} we get 8 possible states:
\[\begin{array}{l}
(\mbox{|e|sta |sel|va |sel|vag|gia e |a|spra e},.648,9,.2)\\
(\mbox{|e|sta |sel|va |sel|vag|gia e a|spra e},.162,8,1)\\
(\mbox{|e|sta |sel|va |sel|vag|gia e |a|spra |e},.072,10,.2)\\
(\mbox{|e|sta |sel|va |sel|vag|gia |e |a|spra e},.072,10,.2)\\
(\mbox{|e|sta |sel|va |sel|vag|gia e a|spra |e},.018,9,.2)\\
(\mbox{|e|sta |sel|va |sel|vag|gia |e a|spra e},.018,9,.2)\\
(\mbox{|e|sta |sel|va |sel|vag|gia |e |a|spra |e},.008,11,.2)\\
(\mbox{|e|sta |sel|va |sel|vag|gia |e a|spra |e},.002,10,.2)\\
\end{array}
\]
The processing of the last word \say{forte} does not introduce additional
non determinism, so we end up with the 8 possible states listed below. 
However, only 3 of them are formed by 11 syllables, while the others 
(emphasized in red) are not admissible:
\[\begin{array}{l}
(\mbox{|e|sta |sel|va |sel|vag|gia e |a|spra e |for|te},.648,11,.2)\\
{\color{red}(\mbox{|e|sta |sel|va |sel|vag|gia e a|spra e |for|te},.162,10,1)}\\
{\color{red}(\mbox{|e|sta |sel|va |sel|vag|gia e |a|spra |e |for|te},.072,12,.2)}\\
{\color{red}(\mbox{|e|sta |sel|va |sel|vag|gia |e |a|spra e |for|te},.072,12,.2)}\\
(\mbox{|e|sta |sel|va |sel|vag|gia e a|spra |e |for|te},.018,11,.2)\\
(\mbox{|e|sta |sel|va |sel|vag|gia |e a|spra e |for|te},.018,11,.2)\\
{\color{red}(\mbox{|e|sta |sel|va |sel|vag|gia |e |a|spra |e |for|te},.008,13,.2)}\\
{\color{red}(\mbox{|e|sta |sel|va |sel|vag|gia |e a|spra |e |for|te},.002,12,.2)}\\
\end{array}
\]
Among the remaining possibilities with eleven syllables we choose the most likely, that is
\begin{center}
    |e|sta |sel|va |sel|vag|gia e |a|spra e |for|te
\end{center}
This choice is consistent with the prosody of Italian language, considering both the strong propensity of \say{e} to left synalephe (“selvaggia e”, “aspra e”), and the propensity to natural hiatus between monosyllables and accented vowel (“e aspra”).

\subsection{Management of Metric constraints}
\label{sec:constraints}
For the reasons we already explained, instead of relying on the total number 
of syllables, it is better to rely on metric constraints. The most important
one is to have a stress on the 10th syllable: if a syllabification does not 
satisfy this constraint it can be pruned. Additionally, we (currently) consider 
the constraint of having either a stress on the 6th or on the 4th syllable.
More sophisticated constraints can be easily integrated in the algorithm. 

In order to avoid to reprocess the entire verse, we add
to each state a small list of boolean flags a4,a6,a10 
expressing the presence of an accent at the corresponding
position. The flag is initialized to False, and set to True
if we find an accent on the expected syllable. 
In the current version, we accept accents 
independently from the grammatical category of the word 
(an information that we do not have in our vocabulary yet).
We only reserve a special treatment to some monosyllables,
mostly articles and prepositions, that we do not accept 
as legal accents.

The constraint on the tenth syllable is mandatory; for the
other metric constraints, the algorithm privileges syllabifications satisfying either the stress in 4th or 
the stress in 6th, raising a warning if none of them is found.
The verses raising a warning are just a dozen, and are 
commented in Appendix\ref{appendix:anomalous}; none of them is 
problematic from the point of view of the syllabification.

\section{Amendments to the Gutenberg edition}
\label{sec:amendments}

The verse\smallskip\\
\begin{tabular}{p{1cm}p{7cm}p{4cm}}
&e suol di state talor essere grama.  &Inferno XX, 81
\end{tabular}\smallskip\\
contains 12 syllables.
Following Petrocchi \cite{Petrocchi}, we amend it as follows:\smallskip\\
\begin{tabular}{p{1cm}p{7cm}p{4cm}}
&e suol di state talor esser grama.  &Inferno XX, 81
\end{tabular}

It is interesting to observe that the previous mistake is common to 
several on line versions (at the time of submission), comprising e.g. 
\href{https://divinacommedia.weebly.com/inferno-canto-xx.html}{https://divinacommedia.weebly.com/inferno-canto-xx.html}, or \href{https://www.hs-augsburg.de/~harsch/italica/Cronologia/secolo14/Dante/dan\_d120.html}{https://www.hs-augsburg.de/~harsch/italica/Cronologia/secolo14/Dante/dan\_d120.html} and many others.

\subsection{Minor modifications}

In the verse,\smallskip\\
\begin{tabular}{p{1cm}p{7cm}p{4cm}}
&Tesëo combatter co’ doppi petti;  &Purgatorio XXIV, 123
\end{tabular}\smallskip\\
we added a tonic accent on \say{combattér} (short for \say{combatterono}), to 
distinguish it from the infinite form \say{combàtter} in e.g.\smallskip\\
\begin{tabular}{p{1cm}p{7cm}p{4cm}}
&licenza di combatter per lo seme  &Paradiso XII, 95
\end{tabular}\smallskip\\

In the verse,\smallskip\\
\begin{tabular}{p{1cm}p{7cm}p{4cm}}
&ch’io drizzava spesso il viso in vano. &Purgatorio IX, 84
\end{tabular}\smallskip\\
\say{ch’io} must be split into two syllables, while it is usually a single one.
By analogy with similar situations in the document, we added a diereris: \say{ch’ïo}.

\section{Problematic cases}
\label{sec:problems}
There still remain a few problematic (and debated) cases. 
The source of complexity is well described by Beccaria \cite{cesura} in his discussion of the \say{dialephe} entry for the Enciclopedia Dantesca Treccani
(translation by the authors):
\begin{quote}
    Indeed, Dante felt deeply in the verse, especially in the Comedy, 
    the rhythmic and logical values (more than melodic fluidity). The dialephe, even of the less usual type in the Comedy, seems in fact very 
    often determined by the pause of thought [\dots] and special artistic justifications were also found \cite{Casella24} (Casella) for slow narrative scans, underlined by dialephe before conjunctions.
\end{quote}
It is remarkable to observe that most of the examples cited by Beccaria are automatically handled
by our approach. For instance in the verse\smallskip

\begin{tabular}{p{1cm}p{7cm}p{4cm}}
&d’infanti e di femmine e di viri &Inferno IV, 30
\end{tabular}\smallskip\\
the dialephe after the word \say{infanti} (children) is imposed by the requirement of a stress 
on the $6^{th}$ syllable. 

Similarly, in 

\begin{tabular}{p{1cm}p{7cm}p{4cm}}
&era già grande, e già eran tratti  &Paradiso XVI, 107
\end{tabular}\smallskip\\
the dialephe after \say{grande} (big) is required in order to have a stress on the $10^{th}$ syllable. 
The most problematic cases are
related to the heuristic rule of {\em detaching hemistics starting with an unstressed vowel in slow-scanned verses dealing with enumerations} \cite{cesura}. 

We just mention the couple of cases that, according to this rule, are incorrectly classified
by our algorithm (as far as we have been able to check); for these verses, we give the 
syllabification produced by our algorithm (former) and the \say{canonical} one (latter):\smallskip

\begin{tabular}{p{1cm}p{7cm}p{4cm}}
&ma |sa|pï|en|za, a|mo|re| e |vir|tu|te &Inferno I, 104\\
&ma |sa|pï|en|za, |a|mo|re e |vir|tu|te &
\end{tabular}\smallskip\\

\begin{tabular}{p{1cm}p{7cm}p{4cm}}
&leg|ge, |mo|ne|ta, of|fi|cio |e |co|stu|me  & Purgatorio VI, 146\\
&leg|ge, |mo|ne|ta, |of|fi|cio e |co|stu|me  &
\end{tabular}\smallskip\\

Dealing with these remaining cases is an interesting challenge
for future developments of our works. However, it is also 
worth to remark that it is precisely the somewhat anomalous 
behaviour of these verses that make them interesting from a linguistic point of view. So, the difference between the intended syllabification and the most likely one produced by our algorithm is {\em per se} an interesting phenomenon, worth to be observed and investigated more than simply corrected.

\section{Conclusions}
\label{sec:conclusion}
We give, for the first time, a complete syllabification of the Divine 
Comedy by Dante Alighieri. The interest of the work from the cultural
heritage point of view is, in our opinion, evident: we enrich the source document with an information that can be directly processed in a lot of 
different and innovative ways, and for different purposes. 
We enrich data, do not provide processing frameworks or applications. 

Our syllabification algorithm, based on techniques borrowed from 
probabilistic and constraint programming, is simple and effective. 
It is not meant to mimic or reflect 
correct metric or phonetic rules: many of ours \say{admissible} syllabifications
are totally incorrect from a metric point of view: we simply rule them
out as unlikely. Our probabilistic approach could actually open new
perspectives about the digital investigation of phonetic rules, and their
empirical justification, especially in view of automatic learning.

While our work has been focused on the Divina Commedia, due to the world-wide
interest on this document, the overall approach may be extended to most 
of the Italian poetic literature. The most delicate issue is the extension
of the dictionary; in particular, it is not evident if the synalephe 
propensities generalize to different authors, and to what extent (that seems
to be by itself a very interesting topic). 

There are a few limitations of our approach, that can be possibly addressed in future works. 
A weak point, that could cause problems in some situations, is
that the synalephe probabilities associated with words are {\em context 
independent}. Another interesting perspective, partially related to the
previous point, consists in taking into account more sophisticated 
phonetic categories (semivowels, approximants, \dots). The management of
punctuation symbols could be revisited in view of the issues in Section~\ref{sec:problems}.
Finally, as discussed in Section \ref{sec:constraints}, the identification of the rhythmic
cadence of a verse could take advantage of the grammatical category of words, that is an 
information currently missing from the vocabulary.



\appendix
\newpage
\section{Syllabification of Inferno, Canto I}
\label{app:example}
We recall our conventions:
\begin{itemize}
    \item group of vowels in the final word of the verse are treated as diphthongs (tro|vai, in|trai, ab|ban|do|nai, \dots)
    \item apostrophes are assimilated to vowels: (on|d'io, do|v'or, \dots)
\end{itemize}\smallskip\smallskip
\begin{center}

\begin{tabular}{r@{\hskip 1cm}l}
  1 &|Nel |mez|zo |del |cam|min |di |no|stra |vi|ta\\             
  2 &|mi |ri|tro|vai |per |u|na |sel|va o|scu|ra,\\               
  3 &|ché |la |di|rit|ta |via |e|ra |smar|ri|ta. \\\\               

  4 &|Ahi |quan|to a |dir |qual |e|ra è |co|sa |du|ra\\           
  5 &|e|sta |sel|va |sel|vag|gia e |a|spra e |for|te\\            
  6 &|che |nel |pen|sier |ri|no|va |la |pa|u|ra!\\\\                

  7 &|Tan|t’ è a|ma|ra |che |po|co |è |più |mor|te;\\             
  8 &|ma |per |trat|tar |del |ben |ch’ i’ |vi |tro|vai,\\         
  9 &|di|rò |de |l’ al|tre |co|se |ch’ i’ |v’ ho |scor|te.\\\\      

 10 &|Io |non |so |ben |ri|dir |com’ |i’ |v’ in|trai, \\          
 11 &|tan|t’ e|ra |pien |di |son|no |a |quel |pun|to \\           
 12 &|che |la |ve|ra|ce |via |ab|ban|do|nai.\\\\                    

 13 &|Ma |poi |ch’ i’ |fui |al |piè |d’ un |col|le |giun|to,  \\  
 14 &|là |do|ve |ter|mi|na|va |quel|la |val|le  \\                
 15 &|che |m’ a|vea |di |pa|u|ra il |cor |com|pun|to,\\\\           

 16 &|guar|dai |in |al|to e |vi|di |le |sue |spal|le \\           
 17 &|ve|sti|te |già |de’ |rag|gi |del |pia|ne|ta  \\             
 18 &|che |me|na |drit|to al|trui |per |o|gne |cal|le.\\\\          

 19 &|Al|lor |fu |la |pa|u|ra un |po|co |que|ta,  \\              
 20 &|che |nel |la|go |del |cor |m’ e|ra |du|ra|ta \\             
 21 &|la |not|te |ch’ i’ |pas|sai |con |tan|ta |pie|ta. \\\\        

 22 &|E |co|me |quei |che |con |le|na af|fan|na|ta,\\             
 23 &|u|sci|to |fuor |del |pe|la|go a |la |ri|va,  \\             
 24 &|si |vol|ge a |l’ ac|qua |pe|ri|glio|sa e |gua|ta, \\\\        

 25 &|co|sì |l’ a|ni|mo |mio, |ch’ an|cor |fug|gi|va,\\           
 26 &|si |vol|se a |re|tro a |ri|mi|rar |lo |pas|so \\            
 27 &|che |non |la|sciò |già |mai |per|so|na |vi|va.\\\\            
 28 &|Poi |ch’ èi |po|sa|to un |po|co il |cor|po |las|so,\\       
 29 &|ri|pre|si |via |per |la |piag|gia |di|ser|ta, \\            
 30 &|sì |che ’l |piè |fer|mo |sem|pre e|ra ’l |più |bas|so.
 \end{tabular}
\newpage

 \begin{tabular}{r@{\hskip 1cm}l}

 31 &|Ed |ec|co, |qua|si al |co|min|ciar |de |l’ er|ta, \\        
 32 &|u|na |lon|za |leg|ge|ra e |pre|sta |mol|to, \\              
 33 &|che |di |pel |ma|co|la|to e|ra |co|ver|ta; \\\\ 
 
 34 &|e |non |mi |si |par|tia |di|nan|zi al |vol|to, \\            
 35 &|an|zi ’m|pe|di|va |tan|to il |mio |cam|mi|no, \\            
 36 &|ch’ i’ |fui |per |ri|tor|nar |più |vol|te |vòl|to.\\\\  
 
 37 &|Tem|p’ e|ra |dal |prin|ci|pio |del |mat|ti|no,  \\          
 38 &|e ’l |sol |mon|ta|va ’n |sù |con |quel|le |stel|le  \\      
 39 &|ch’ e|ran |con |lui |quan|do |l’ a|mor |di|vi|no\\\\          

 40 &|mos|se |di |pri|ma |quel|le |co|se |bel|le; \\              
 41 &|sì |ch’ a |be|ne |spe|rar |m’ e|ra |ca|gio|ne \\            
 42 &|di |quel|la |fie|ra a |la |ga|et|ta |pel|le  \\\\             

 43 &|l’ o|ra |del |tem|po e |la |dol|ce |sta|gio|ne;  \\         
 44 &|ma |non |sì |che |pa|u|ra |non |mi |des|se \\               
 45 &|la |vi|sta |che |m’ ap|par|ve |d’ un |le|o|ne. \\\\           

 46 &|Que|sti |pa|rea |che |con|tra |me |ve|nis|se  \\            
 47 &|con |la |te|st’ al|ta e |con |rab|bio|sa |fa|me, \\         
 48 &|sì |che |pa|rea |che |l’ ae|re |ne |tre|mes|se.\\\\           

 49 &|Ed |u|na |lu|pa, |che |di |tut|te |bra|me   \\              
 50 &|sem|bia|va |car|ca |ne |la |sua |ma|grez|za, \\             
 51 &|e |mol|te |gen|ti |fé |già |vi|ver |gra|me, \\\\              

 52 &|que|sta |mi |por|se |tan|to |di |gra|vez|za  \\             
 53 &|con |la |pa|u|ra |ch’ u|scia |di |sua |vi|sta, \\           
 54 &|ch’ io |per|dei |la |spe|ran|za |de |l’ al|tez|za.\\\\        

 55 &|E |qual |è |quei |che |vo|lon|tie|ri ac|qui|sta, \\         
 56 &|e |giu|gne ’l |tem|po |che |per|der |lo |fa|ce, \\          
 57 &|che ’n |tut|ti |suoi |pen|sier |pian|ge e |s’ at|tri|sta; \\\\

 58 &|tal |mi |fe|ce |la |be|stia |san|za |pa|ce,\\               
 59 &|che, |ve|nen|do|mi ’n|con|tro, a |po|co a |po|co \\         
 60 &|mi |ri|pi|gne|va |là |do|ve ’l |sol |ta|ce. \\\\              

 61 &|Men|tre |ch’ i’ |ro|vi|na|va in |bas|so |lo|co,\\           
 62 &|di|nan|zi a |li oc|chi |mi |si |fu |of|fer|to \\            
 63 &|chi |per |lun|go |si|len|zio |pa|rea |fio|co.       
\end{tabular}
 
 \newpage
 \begin{tabular}{r@{\hskip 1cm}l}
  64 &|Quan|do |vi|di |co|stui |nel |gran |di|ser|to,\\            
 65 &« |Mi|se|re|re |di |me», |gri|dai |a |lui,  \\                
 66 &« |qual |che |tu |sii, |od |om|bra od |o|mo |cer|to!».\\\\   
 
 67 &|Ri|spuo|se|mi:« |Non |o|mo, o|mo |già |fui, \\              
 68 &|e |li |pa|ren|ti |miei |fu|ron |lom|bar|di, \\              
 69 &|man|to|a|ni |per |pa|trï|a am|be|dui.\\\\       
 
 70 &|Nac|qui |sub |Iu|lio, an|cor |che |fos|se |tar|di, \\       
 71 &|e |vis|si a |Ro|ma |sot|to ’l |buo|no Au|gu|sto \\          
 72 &|nel |tem|po |de |li |dèi |fal|si e |bu|giar|di.\\\\           

 73 &|Po|e|ta |fui, |e |can|tai |di |quel |giu|sto \\             
 74 &|fi|gliuol |d’ An|chi|se |che |ven|ne |di |Tro|ia,\\         
 75 &|poi |che ’l |su|per|bo I|lï|ón |fu |com|bu|sto.\\\\           

 76 &|Ma |tu |per|ché |ri|tor|ni a |tan|ta |no|ia?\\              
 77 &|per|ché |non |sa|li il |di|let|to|so |mon|te\\              
 78 &|ch’ è |prin|ci|pio e |ca|gion |di |tut|ta |gio|ia?». \\\\     

 79 &« |Or |se’ |tu |quel |Vir|gi|lio e |quel|la |fon|te \\        
 80 &|che |span|di |di |par|lar |sì |lar|go |fiu|me?», \\         
 81 &|ri|spuo|s’ io |lui |con |ver|go|gno|sa |fron|te.\\\\          

 82 &« |O |de |li al|tri |po|e|ti o|no|re e |lu|me, \\             
 83 &|va|glia|mi ’l |lun|go |stu|dio e ’l |gran|de a|mo|re \\     
 84 &|che |m’ ha |fat|to |cer|car |lo |tuo |vo|lu|me. \\\\          

 85 &|Tu |se’ |lo |mio |ma|e|stro e ’l |mio |au|to|re,\\          
 86 &|tu |se’ |so|lo |co|lui |da |cu’ |io |tol|si \\              
 87 &|lo |bel|lo |sti|lo |che |m’ ha |fat|to o|no|re. \\\\          

 88 &|Ve|di |la |be|stia |per |cu’ |io |mi |vol|si; \\            
 89 &|a|iu|ta|mi |da |lei, |fa|mo|so |sag|gio,  \\                
 90 &|ch’ el|la |mi |fa |tre|mar |le |ve|ne e i |pol|si».\\\\       

 91 &« |A |te |con|vien |te|ne|re al|tro |vï|ag|gio», \\           
 92 &|ri|spuo|se, |poi |che |la|gri|mar |mi |vi|de,  \\           
 93 &« |se |vuo’ |cam|par |d’ e|sto |lo|co |sel|vag|gio;\\\\         

 94 &|ché |que|sta |be|stia, |per |la |qual |tu |gri|de,\\        
 95 &|non |la|scia al|trui |pas|sar |per |la |sua |via,\\         
 96 &|ma |tan|to |lo ’m|pe|di|sce |che |l’ uc|ci|de; \\\\                
\end{tabular}
 \newpage
 
\begin{tabular}{r@{\hskip 1cm}l}
 97 &|e |ha |na|tu|ra |sì |mal|va|gia e |ria,\\                   
 98 &|che |mai |non |em|pie |la |bra|mo|sa |vo|glia,\\            
 99 &|e |do|po ’l |pa|sto ha |più |fa|me |che |pria. \\\\  
 
100 &|Mol|ti |son |li a|ni|ma|li a |cui |s’ am|mo|glia, \\        
101 &|e |più |sa|ran|no an|co|ra, in|fin |che ’l |vel|tro\\       
102 &|ver|rà, |che |la |fa|rà |mo|rir |con |do|glia.\\\\            
103 &|Que|sti |non |ci|be|rà |ter|ra |né |pel|tro, \\             
104 &|ma |sa|pï|en|za, a|mo|re |e |vir|tu|te,  \\                 
105 &|e |sua |na|zion |sa|rà |tra |fel|tro e |fel|tro. \\\\ 

106 &|Di |quel|la u|mi|le I|ta|lia |fia |sa|lu|te \\              
107 &|per |cui |mo|rì |la |ver|gi|ne |Cam|mil|la, \\               
108 &|Eu|ria|lo e |Tur|no e |Ni|so |di |fe|ru|te. \\\\              

109 &|Que|sti |la |cac|ce|rà |per |o|gne |vil|la, \\              
110 &|fin |che |l’ av|rà |ri|mes|sa |ne |lo ’n|fer|no,\\          
111 &|là |on|de ’n|vi|dia |pri|ma |di|par|til|la. \\\\              

112 &|On|d’ io |per |lo |tuo |me’ |pen|so e |di|scer|no \\        
113 &|che |tu |mi |se|gui, e |io |sa|rò |tua |gui|da,\\           
114 &|e |trar|rot|ti |di |qui |per |lo|co et|ter|no; \\\\           

115 &|o|ve u|di|rai |le |di|spe|ra|te |stri|da, \\                
116 &|ve|drai |li an|ti|chi |spi|ri|ti |do|len|ti, \\             
117 &|ch’ a |la |se|con|da |mor|te |cia|scun |gri|da;\\\\           

118 &|e |ve|de|rai |co|lor |che |son |con|ten|ti \\               
119 &|nel |fo|co, |per|ché |spe|ran |di |ve|ni|re \\              
120 &|quan|do |che |sia |a |le |be|a|te |gen|ti.\\\\                

121 &|A |le |quai |poi |se |tu |vor|rai |sa|li|re,\\              
122 &|a|ni|ma |fia |a |ciò |più |di |me |de|gna:  \\              
123 &|con |lei |ti |la|sce|rò |nel |mio |par|ti|re; \\\\        

124 &|ché |quel|lo im|pe|ra|dor |che |là |sù |re|gna, \\          
125 &|per|ch’ i’ |fu’ |ri|bel|lan|te a |la |sua |leg|ge, \\       
126 &|non |vuol |che ’n |sua |cit|tà |per |me |si |ve|gna.\\\\      

127 &|In |tut|te |par|ti im|pe|ra e |qui|vi |reg|ge; \\           
128 &|qui|vi è |la |sua |cit|tà |e |l’ al|to |seg|gio:\\          
129 &|oh |fe|li|ce |co|lui |cu’ |i|vi e|leg|ge!».           

\end{tabular}

\newpage

\begin{tabular}{r@{\hskip 1cm}l}
130 &|E |io |a |lui:« |Po|e|ta, io |ti |ri|cheg|gio   \\          
131 &|per |quel|lo |Dio |che |tu |non |co|no|sce|sti,  \\         
132 &|ac|ciò |ch’ io |fug|ga |que|sto |ma|le e |peg|gio,\\\\        

133 &|che |tu |mi |me|ni |là |do|v’ or |di|ce|sti,\\              
134 &|sì |ch’ io |veg|gia |la |por|ta |di |san |Pie|tro  \\       
135 &|e |co|lor |cui |tu |fai |co|tan|to |me|sti».\\\\              

136 &|Al|lor |si |mos|se, e |io |li |ten|ni |die|tro.   
\end{tabular}
\end{center}
\bigskip

\section{Hiatuses and Diphthongs in the Divine Comedy}
\label{appendix:hiatuses}

In this appendix, we provide a more exhaustive discussion about the syllabification of words in the Divine Comedy. The main purpose
of this Section
is to highlight the complexity of the problem and the differences 
with respect to the syllabification of modern Italian. The most interesting cases have been anticipated in Section~\ref{sec:word-syllabification};
here we discuss a few additional issues regarding the use of
hiatuses and diphthongs in the Divine Comedy. We organize the discussion
around the most problematic combination of vowels: ia, ie, io, ea.

From a linguistic point of view, our discussion may actually appear over simplistic; actually, the presentation is mostly given from a computational perspective, and influenced by the current version
of the algorithm. In particular, at the moment, we have no 
information about the position of the accent in a combination of vowels (this could be taken into account in future versions). In fact, in Italian poetry (and natural language) we can distinguish three situations: 
\begin{enumerate}
\item when the accent falls on the second vowel (ascendent nexus, es. patri-àrca, ri-òne, disi-àre) normally we have a hiatus (but never in the frequent cases of words coming in Florentine from the Latin PL: pieno from PLENUS, and in some other circumstances of real diphthongs \cite{Menichetti93}); 
\item when the accent falls on the first vowel (descendent nexus, es. farìa, ìo, mìo, arpìa, corsìa) the nexus is more often monosyllabic; 
\item when both vowels are out of accent (unstressed nexus, es. accìdia, ambròsia, viaggiàre, trionfàre) the situation is far more complicated, but in poetry prevails the hiatus \cite{Menichetti93}. 
\end{enumerate}

\subsection{i-a}
The vowel \say{i} followed by \say{a}, that typically constitutes a 
diphthong
in modern Italian, is frequently a hiatus in the Divine Comedy (especially in the middle of words). 
A couple of frequent and relatively simple cases are the following:
\begin{description}
\item[viaggio]\,

\begin{tabular}{p{1cm}p{6cm}p{4cm}}
&\say{A te convien tenere altro viaggio}, &Inferno, I, 91
\end{tabular}\smallskip\\
Other occurrences are in Inferno, X, 132; Inferno XVI, 27; Inferno XXVIII, 16; Inferno XXXI, 82;
Purgatorio, II, 92.\smallskip

\item[disiata] similarly to \say{disiato}, \say{disiando}, \say{disiar},\say{disianza}, etc.\smallskip\\
\begin{tabular}{p{1cm}p{6cm}p{4cm}}
&Vostra parola disiata vola, &Purgatorio I, 83
\end{tabular}\smallskip\\
Other occurrences are in Inferno V, 133; Inferno XXX, 140, Purgatorio III, 40, Purgatorio XXIX, 5; Purgatorio XXIX, 33; Purgatorio XXXIII, 83; Paradiso III, 73; , Paradiso V, 86; Paradiso XV, 66;; Paradiso XXII, 18; Paradiso XXII, 65; Paradiso XXIII, 4; Paradiso XXIII, 14; Paradiso XXIII, 39; Paradiso XXX, 15;

\end{description}
Other words that contain a ia-hiatus are: 

accidia, ambrosia, Anfiarao, anzian, Briareo, eresiarche, celestiali, Ciriatto, clementiae, convertian, disia, disviando, disviato, Fialte, 
giovial, gloriar, gloriarla,
gratia, Grazian, inebriate, India, inebriate, inebriava, ’nvetriate, inviasti, Iustiniano, Labia, lilia, Madian, mandrian, Mariza, Marsia,
meridian, meridiana, Oriaco, Polinmia, potenziata, radial, radiando, Rialto, scienzia, scuriada, spezial, storiata, straniasse, sustanzial, svia, sviando, sviati, triangol, Trivia, umiliato, variar, variazion, venian, Vitaliano, Zodiaco 

More complex is the case of the following words, sometimes pronounced with a hiatus, and sometimes with a diphthong. For each word, we provide an example
of the two possibilities (we use a dieresis to remark the difference)\smallskip

fiate,fiata

\begin{tabular}{p{1cm}p{6cm}p{4cm}}
 &spesse fïate ragioniam del monte  &Purgatorio XXII, 104\\\smallskip
 &se mille fiate in sul capo mi tomi.  &Inferno XXXII, 102\\
\end{tabular}\smallskip

patria:

\begin{tabular}{p{1cm}p{6cm}p{4cm}}
 &di quella nobil patrïa natio,  &Inferno X, 26\\\smallskip
 &e non molto distanti a la tua patria, &Paradiso XXI, 107\\
\end{tabular}\smallskip

%

infamia  

\begin{tabular}{p{1cm}p{6cm}p{4cm}}
 &l’infamïa di Creti era distesa  &Inferno XII, 12\\\smallskip
 &sanza tema d’infamia ti rispondo.  &Inferno XXVII, 66\\
\end{tabular}\smallskip

venian

\begin{tabular}{p{1cm}p{6cm}p{4cm}}
 &e non pareva, sì venïan lente.  &Purgatorio III, 60\\\smallskip
 &che venian lungo l’argine, e ciascuna  &Inferno XV, 17\\
\end{tabular}\smallskip

celestial

\begin{tabular}{p{1cm}p{6cm}p{4cm}}
 &celestïal giacer, da l’altra parte,  &Purgatorio XII, 29\\\smallskip
 &Da poppa stava il celestial nocchiero,  &Purgatorio II, 43
\end{tabular}\smallskip

gloria 

\begin{tabular}{p{1cm}p{6cm}p{4cm}}
 &‘Glorïa in excelsis’ tutti ‘Deo’  &Purgatorio XX, 136\\\smallskip
 &de la mia gloria e del mio paradiso &Paradiso XV, 36
\end{tabular}\smallskip

Although the sequence i-a frequently constitutes a hiatus, this is not a norm. In the frequent case in which the accent is on the second vowel,
we have a natural diphthong. Examples, just borrowed from the first canto, are: pianeta, piaggia, piange, bestia, Troia, noia, gioia, Italia.

\subsection{i-e}
The two vowels \say{ie} form a hiatus in the following words:

Ariete, audienza, balbuziendo, coscienza, Daniel, Daniello, dieta, esperienza, esuriendo, Ezechiel, Gabriel, Gabriello, Galieno,  ’nvieranno, niente, odierno, oriental, oriente, Ostiense, pazienza, pietate, progenie, quieta, quietar, quietarmi, quietata, quiete, quieto, quietò, requievi, riempion, riesca, sapienza, scienza, scienzia, Siestri . 

The two vowels \say{ie} may be a hiatus in the following words:\smallskip

pieta/pietate 

\begin{tabular}{p{1cm}p{6cm}p{4cm}}
 &con buona pïetate aiuta il mio!  &Purgatorio V, 87\\\smallskip
 &In te misericordia, in te pietate,  &Paradiso XXXIII, 19
\end{tabular}\smallskip

obediendo/disobediendo

\begin{tabular}{p{1cm}p{6cm}p{4cm}}
 &con umiltate obedïendo poi &Paradiso VII, 99\\\smallskip
 &quanto disobediendo intese ir suso;  &Paradiso VII, 100
\end{tabular}\smallskip

sufficiente/sufficienti

\begin{tabular}{p{1cm}p{6cm}p{4cm}}
 &acciò che re sufficïente fosse;  &Paradiso XIII, 96\\\smallskip
 &per far l’uom sufficiente a rilevarsi  &Paradiso VII, 116
\end{tabular}\smallskip

Examples of natural dipththongs from the first Canto are:
pensier, pien, fiera, volontieri, miei, convien, empie, Pietro, dietro.

\subsection{i-o}
The two vowels \say{io} are a hiatus in the following words:

accidioso, Anfione, anterior, aspersion, Caliopè, Curio, 
Diogenès, Diomede, Dione, Dionisio, disio, disioso, division,
elezion, elezione, elezioni, Eliodoro,
Etiopia, Etiopo, furiosa, 
gaudiose, gaudioso, Gerion, Gerione, gloriosa, gloriosamente, glorioso, gloriosi, idioma, Iperione,lioncel, Livio, lussuriosa, Niobè, 
oblivion, oppinion, oppinione, Pigmalione, pio,
piorno, presunzion, preziosa, prezioso,
razionabile, region, religione,
Scariotto, Scipion, Scipione, settentrion, settentrional, studiose, violenta, violenti, viole, violenza, vision, visione,

The two vowels \say{io} may be a hiatus in the following words:\smallskip

conversione

\begin{tabular}{p{1cm}p{6cm}p{4cm}}
&La mia conversïone, omè!, fu tarda;  &Purgatorio XIX, 106\\\smallskip
&e per trovare a conversione acerba  &Paradiso XI, 103
\end{tabular}\smallskip

distinzione

\begin{tabular}{p{1cm}p{6cm}p{4cm}}
&sanza distinzïone in essordire.  &Paradiso XXIX, 30\\\smallskip
&che sanza distinzione afferma e nega  &Paradiso XIII, 116
\end{tabular}\smallskip

grazioso

\begin{tabular}{p{1cm}p{6cm}p{4cm}}
&O animal grazïoso e benigno  &Inferno V, 88\\\smallskip
&ditemi, ché mi fia grazioso e caro,  &Purgatorio XIII, 91
\end{tabular}\smallskip

invidiosa/invidiosi

\begin{tabular}{p{1cm}p{6cm}p{4cm}}
&silogizzò invidïosi veri.  &Paradiso X, 138\\\smallskip
&gent’ è avara, invidiosa e superba:  &Inferno XV, 68
\end{tabular}\smallskip

orazion/orazione

\begin{tabular}{p{1cm}p{6cm}p{4cm}}
&se buona orazïon lui non aita,  &Purgatorio XI, 130\\\smallskip
&così, a l’orazion pronta e divota,  &Paradiso XIV, 22
\end{tabular}\smallskip

passion

\begin{tabular}{p{1cm}p{6cm}p{4cm}}
&quand’ ira o altra passïon ti tocca!  &Inferno XXXI, 72\\\smallskip
&a la passion di che ciascun si spicca,  & Purgatorio XXI, 107
\end{tabular}\smallskip

perfezion/perfezione

\begin{tabular}{p{1cm}p{6cm}p{4cm}}
&di tutta l’animal perfezïone;  &Paradiso XIII, 83\\\smallskip
&sanza sua perfezion fosser cotanto.  &Paradiso XXIX, 45
\end{tabular}\smallskip

The cases of the two words \say{mio} (my) and \say{io} (I) possibly deserves a 
little discussion.  

The word \say{mio} occurs 310 times in the Divine Comedy, and in 307 cases
it must be read as a single syllable. Let us briefly review the remaining three cases, namely

\begin{tabular}{p{1cm}p{6cm}p{4cm}}
&ma quella folgorò nel mïo sguardo  &Paradiso III, 128\smallskip\\
&Tal vero a l’intelletto mïo sterne  &Paradiso XXVI, 37\smallskip\\
&già tutta mïo sguardo avea compresa, &Paradiso XXXI, 53\smallskip\\
\end{tabular}\smallskip\\
These verses are discussed in \cite{Menichetti93}, p.250, who remarks
the relation with the spurious s- following \say{mïo}. Actually, the
ancient Florentine language still maintains the {\em prostesi} in many cases (\say{iscritto}, \say{isguardo}, \dots); the potential {\em prostesi},
helps the separation of the previous nexus.

%
%

%

The case of \say{io} (I) is even more problematic. 
In the Gutenberg edition, we have 28 occurrences of \say{ïo}, versus 679 occurrences of \say{io}.  
Due to the synalephe,
in most of the cases the two forms \say{io/ïo} would not change neither 
the total number of syllables, nor the position of tonic stresses inside 
the verse. The hiatus \say{ïo} may look closer to the Latin etymology \say{ego},
but in the frequent cases when synalephe is impossible, it is clear 
that \say{io} normally constitutes a single syllable. Some examples are:

\begin{tabular}{p{1cm}p{6cm}p{4cm}}
&rispuos’ io lui con vergognosa fronte.  &Inferno I, 81\\\smallskip
&tu se’ solo colui da cu’ io tolsi  &Inferno I, 86\\\smallskip
&Io non posso ritrar di tutti a pieno,  &Inferno IV, 145\\\smallskip
&e di questi cotai son io medesmo.  & Inferno IV, 39\\\smallskip
&Io venni in loco d’ogne luce muto,  &Inferno V, 28
\end{tabular}

Nevertheless, there are situations where a hiatus is imposed for metric
reasons. A nice example is the following verse, where the two forms 
coexists:

\begin{tabular}{p{1cm}p{6cm}p{4cm}}
&Cred’ ïo ch’ei credette ch’io credesse  &Inferno XIII, 25
\end{tabular}

Let us also observe that shifting the hiatus on the second occurrence
of \say{io} would not change the total number of syllables. However, 
the hiatus on the first occurrence allows us to have a stress on the $6th$
syllable, in \say{credètte}, and is eventually correct. This is a clear indication
that for the correct syllabification of verses relying on the stress 
in $10th$ is not enough.

The linguistic motivation to privilege the dieresis on the first occurrence of “io” is that, coming after its verb, it is under stress and produces a more natural dieresis \cite{Menichetti93}.

%

\subsection{e-a}
\label{appendix:ea}
The group \say{ea} is the one where the use of dieresis in the Gutenberg 
edition is more questionable. 

The word \say{ideale}
is expressed with dieresis, but this looks somehow redundant.


Apart from a few proper names (Beatrice, Enea, Rea, Rodopea, Tarpea, \dots)
the remaining cases are verbs at the imperfect: 

avea, avean, correan, dicea, discendea,
discernea, dovea, facea, facean, giacea, imprendea, intendea,  
parea, parean, percoteansi, piacea, piangea, potea, potean,
ravvolgea, reflettea, sapean, solea, tenea,
vedea, vincea.

In all these verbs, \say{ea} may constitute both a hiatus or a diphthong (where the latter is, by far, the norm).
We give a few examples

avea

\begin{tabular}{p{1cm}p{6cm}p{4cm}}
&fiso nel punto che m’avëa vinto.  &Paradiso XXIX, 9\\\smallskip
&ma l’un de’ cigli un colpo avea diviso.  &Purgatorio III, 108
\end{tabular}\smallskip

avean

\begin{tabular}{p{1cm}p{6cm}p{4cm}}
&udito avëan l’ultimo costrutto;  &Purgatorio XXVIII, 147\\\smallskip
&Non avean penne, ma di vispistrello  &Inferno XXXIV, 49
\end{tabular}\smallskip

correa/correan

\begin{tabular}{p{1cm}p{6cm}p{4cm}}
&corrëan genti nude e spaventate,  &Inferno XXIV, 92\\\smallskip
&e correa contro ’l ciel per quelle strade  &Purgatorio XVIII, 79
\end{tabular}\smallskip

dovea

\begin{tabular}{p{1cm}p{6cm}p{4cm}}
&Sì com’ io fui, com’ io dovëa, seco,  &Purgatorio XXXIII, 22\\\smallskip
&lo qual dovea Penelopè far lieta,  &Inferno XXVI, 96
\end{tabular}\smallskip

facea

\begin{tabular}{p{1cm}p{6cm}p{4cm}}
&ch’io facëa dinanzi a la risposta,  &Inferno X, 71\\\smallskip
&sì che ’l sangue facea la faccia sozza,  &Inferno XXVIII, 105
\end{tabular}\smallskip

\newpage
parean

\begin{tabular}{p{1cm}p{6cm}p{4cm}}
&e or parëan da la bianca tratte,  &Purgatorio XXIX, 12\\\smallskip
&Morti li morti e i vivi parean vivi:  &Purgatorio XII, 67
\end{tabular}\smallskip

piangea

\begin{tabular}{p{1cm}p{6cm}p{4cm}}
&Io non piangëa, sì dentro impetrai:  &Inferno XXXIII, 49\\\smallskip
&quando piangea, vi facea far le grida.  &Inferno XIV, 102
\end{tabular}\smallskip

However, there are a few cases where Petrocchi's use of dieresis looks somehow abused. Consider for instance the following verses:\smallskip

\begin{tabular}{p{1cm}p{6cm}p{4cm}}
&Ella non ci dicëa alcuna cosa,  &Purgatorio VI,
64\smallskip\\
&che ’l cibo ne solëa essere addotto &Purgatorio XXXIII, 44\smallskip\\
\end{tabular}

As a matter of fact, there is no reason to expect synalephe
in these cases (and hence no reason add a dieresis), precisely {\em because} the words on the left end with a diphthong. 
This is a case of the already mentioned {\em diesinalefe} rule
by Menichetti: when the possibility of a dieresis meets the possibility of a dialephe, the dialephe prevails. 

The words discendea, discernea, giacea, intendea, piacea, ravvolgea, 
reflettea, tenea, vedea and vincea, as well as many occurrences
of other verbs at the imperfect are in a similar situation.

A very interesting case is

\begin{tabular}{p{1cm}p{6cm}p{4cm}}
&che non parëa s’era laico o cherco. &Inferno XVIII, 117\smallskip
\end{tabular}

We accept Petrocchi’s difficult choice to suggest an exceptional dieresis on \say{parëa} because there are no elements (rithmical, linguistical, nor according to Dante’s usus scribendi) to privilege other solutions, equally exceptional:

\begin{tabular}{p{1cm}p{6cm}p{4cm}}
&che |non |pa|rea |s’e|ra |lä|i|co o |cher|co &\smallskip\\
&che |non |pa|rea |s’e|ra |lai|co |o |cher|co &\smallskip
\end{tabular}

Lai|co (normally bisyllabic) could support an exceptional dieresis thanks to the Greek LAÏKÓS, Latin LÄICUS. While parëa can rely on the alternative frequent forms of the imperfect in -eva (pa|re|va).

\section{Anomalous verses}
\label{appendix:anomalous}
There are just a dozen of cases in the Divine Comedy where
the syllabification algorithm is raising a warning due to
the absence of an accent on either the 4th or the 6th syllable.
We have been glad to discover that all of them have been treated and discussed in the secular literary-critical tradition. 

When an adverb in -mente is involved, we can enforce a secondary stress on the first component of the compound word (canìna-mente, miràbil-mente, glorïòsa-mente):

\begin{tabular}{p{1cm}p{6cm}p{4cm}}
&con tre gole caninamente latra  &Inferno VI, 14\smallskip\\
&e vidila mirabilmente oscura.  &Inferno XXI, 6\smallskip\\
&cotanto glorïosamente accolto.  &Paradiso XI, 12\smallskip
\end{tabular}\\
and a similar hypothesis (sustàn-zïàl from sustànza) can be advanced for\vspace{.1cm}

\begin{tabular}{p{1cm}p{6cm}p{4cm}}
&Ogne forma sustanzïal, che setta  &Purgatorio XVIII, 49 \smallskip
\end{tabular}\\
The remaining verses are just considered as hendecasyllables with anomalous accentuation (2nd and 8th, or 3rd and 8th syllable):

\begin{tabular}{p{1cm}p{6cm}p{4cm}}
&mi pinser tra le sepulture a lui, &Inferno X, 38\smallskip\\
&parea che di quel bulicame uscisse &Inferno XII, 117\smallskip\\
&le lagrime, che col bollor diserra,  &Inferno XII, 136\smallskip\\
&per lo furto che frodolente fece  &Inferno XXV, 29\smallskip\\
&la vipera che Melanesi accampa,  &Purgatorio VIII, 80\smallskip\\
&e ‘Beati misericordes!’ fue  &Purgatorio XV, 38\smallskip\\
&e Cesare, per soggiogare Ilerda,  &Purgatorio XVIII, 101\smallskip\\
\end{tabular}\smallskip\\

\end{document}